# A Novel Deep Learning Method for Thermal to Annotated Thermal-Optical Fused Images

Suranjan Goswami, *IEEE* Student Member, Satish Kumar Singh, Senior Member, *IEEE* and Bidyut B. Chaudhuri, Life Fellow, *IEEE*

**Abstract**

*Thermal Images profile the passive radiation of objects and capture them in grayscale images. Such images have a very different distribution of data compared to optical colored images. We present here a work that produces a grayscale thermo-optical fused mask given a thermal input. This is a deep learning based pioneering work since to the best of our knowledge, there exists no other work on thermal-optical grayscale fusion. Our method is also unique in the sense that the deep learning method we are proposing here works on the Discrete Wavelet Transform (DWT) domain instead of the gray level domain. As a part of this work, we also present a new and unique database for obtaining the region of interest in thermal images based on an existing thermal visual paired database, containing the Region of Interest on 5 different classes of data. Finally, we are proposing a simple low cost overhead statistical measure for identifying the region of interest in the fused images, which we call as the Region of Fusion (RoF). Experiments on the database show encouraging results in identifying the region of interest in the fused images. We also show that they can be processed better in the mixed form -rather than with only thermal images.*

## 1. Introduction

Compared to optical images, the thermal Images are difficult to work with because the objects are not well segregated like optical images and different signatures are visible in black and white. This is because Thermal Infrared (TIR) images work on the principle of passive thermal radiation, as opposed to reflected light in optical images or Near Infrared (NIR) images. Moreover, thermal images are produced by radiation, which are of higher wavelength than visible light, leading to a lower resolution. As such, while there exist works like [2-4] which focus on thermal images, we did not come across any work which tries to fuse thermal and optical images represented in the grayscale domain directly. While we have presented a work [5] which tries to prepare color images in a fused domain, that is different from our present work, because we try to create a fused image which can be used in the optical domain. E.g. we can do Deep Learning (DL) based colorization-trained only on optical images, which would not be possible with the work described in [5]. We demonstrate this in Fig. 2. Moreover, our work is specifically focused on data synthesis in a single grayscale level instead of a mask in the RGB or the 3 channel luminance-chrominance (LAB) domain, and thus we have different data distributions for our deep network. We anticipate that our work will be useful in domains that need the input from thermal images to process the data for further information like defense, drone imaging at night, forensic domain images etc.

Also, our method is unique in the sense that we are proposing a network that works on a transformed space (DWT). Our network creates different levels of parallel pathways for DWT transform and captures the data distribution at different levels of abstraction [21]. Finally, we return to the normalized gray-level image domain, the reason for which is described in Section 2.1.

Non DL methods on fusion of thermal images include works like [39] which works on contour formation, [40] for long range observation and [41] which handles multiple thermal image fusion for focus improvement. The DL based methods usually need more data, but in many cases such methods outperform even humans. Some examples are, fine grain image identification [35], image coloring [37, 38] or medical image diagnosis [34, 36]. Thus, in specialized areas, DL based methods are used to handle jobs that are difficult in classical methods. This is one motivation of our current work.

We hypothesize that the distribution of the fused image is similar to both thermal and optical image. Therefore, these images should be properly processed by a machine learning method which is trained to work on either optical or thermal images with nominal retraining. In Sec.3 we show this in our blind testing method. We also provide objective measures in support of our claim in Table 2.

Our database [19] is based on an existing database [6], which contains complex real-world scenes of 5 classes, namely nature, modern infrastructure, animal, human and crowd which we collected over a period of 1.5 years. These images were picked from our work on cross domain colorized images [5], which were not annotated. We manually annotated all collected images and marked the Regions of Interest (ROI). Since these were real (non-synthetic) images, the total process took about 130 working hours to complete. We went on to fuse these images with the input thermal image to obtain the fused image. Each image is then finally Histogram Equalized to obtain the final presented output. A database called CVC-14, [24] has annotated thermal images, but the database has only 1 class annotated namely pedestrians unlike 5 classes in ours. Such annotated databases are not publicly available. Our database provides data distributions which are widely different from each other, which is needed in training DL based models.

We also present a simple new statistical measure for obtaining the region which has been changed most in the output image in comparison to the input image, which we call Region of Fusion (RoF). In summary:

- We demonstrate that it is possible to produce grayscale fused images containing information from both the thermal and optical images.
- We introduce a novel DL architecture that works on a separate logical space (DWT) than the input or output space (normalized images).
- We introduce a unique dataset containing annotated thermal images across multiple classes based on our existing database [6].
- We define a simple statistical score for focusing on a region of interest in fused images.

## 2. Related Works

Machine Learning techniques for working with thermal images have been growing significantly over the last few years. This includes methods for reconstruction of thermal images [46], super resolution [47] imparting color to thermal images [2-5, 37, 42-45], depth estimation [48] and even unsupervised data extraction [49-50]. Similarly, innumerable methods exist for optical domain image processing including for colorization [38, 43], automatic annotation [8-9], denoising [27] etc. However, we have not encountered any work related to processing the TIR images via a fusion method in the grayscale domain. We opted to work on this domain because we hoped to be able to extract and process the information in the fused domain better than either the thermal or the optical domain images individually. We chose Discrete Wavelet Transform as the base of our work as it has been used extensively over the years for processing different kinds of data distributions from audio [26], image compression [31, 27], face detection [17], spliced image detection [32] and even generalized signal [28]. Almost all of these works focused on either restoration or detection. This is primarily because DWT offers an easy to use method which transforms the input signal into a separate (frequency) domain, which helps observe the data from a different viewpoint. This in turn, is used to separate the high frequency and the low frequency information in the data, often even at the pixel level in images [29] providing a statistically inexpensive method.

While there are several works on non DL based image fusion techniques, very few of them like [39-41] handle thermal-optical fusion. All of them work by trying to find an output following some pre-defined rules. CNN tries to achieve this by aiming to find the optimal data distribution [30] for a given input. While there are fusion based DL networks which discriminatively train on multiple domains while providing outputs, like [16], we have not encountered a work that calculates the kernels and computes the full model in a different logical space. We do this because of 2 reasons. The first is that we are eliminating the preprocessing step of converting the visible image into the discrete wavelet transformed data. The second, and more important one is that transforming an image from the visible domain into the DWT domain converts the image into 4 different sub bands for image enhancement. This is often used for blurriness reduction. However, this comes at the cost of distortion of the input images, often at the corners [33]. TIR images possess a data distribution that is blurry by virtue of its capturing sensors and thus, a preprocessing DWT method would result in further degradation of the input data distribution. Hence, we argued that instead of using a discrete wavelet transformed image, if we use a normalized image as the input instead, we should be able to minimize this initial degradation. Of course, an argument can be made that Convolutional Neural Networks (CNNs) themselves work in a separate function space than the input/output space, but we are going beyond that to propose a method that works on a logical space mirroring an established statistical method for signal processing and go on to show that deep networks are capable of effectively learning relations in this space.

Thus, while there are existing works like [21] and [23] that work on the DWT domain, it must be understood that these use wavelet transformed feature maps as input, thus changing the domain of the work completely from visible images. Our method, on the other hand, uses images directly and computes all relations directly in the proposed deep network. Even on non-image based fields with DWT based DL, like in [22], we see that the methods are based on pre-processing the data to obtain the $n^{th}$ level decomposition at the first level before feeding it into the deep network. We avoid this step directly, thus simplifying our process considerably.

## 3. Proposed Method

We are using a deep network to produce a mask, creating a masked image followed by Histogram Equalization to create the final output. The output mask is created by training the model to optimize the loss from an input thermal image to an output image which is the thermal image embedded with the optical-thermal average in the annotated region. This ensures that only a particular region in the thermal image is different from the input image, thus highlighting it. The deep model we are using is described in Fig. 1 and described in Table 1 and our data is presented in Supplementary Section 1.

### 3.1. Deep Network

Our network can roughly be divided into 3 different blocks: Input Encoder/Decoder, DWT Layer and Output Encoder/Decoder. We outline this in Table 1.

Table 1: Deep Network Details

| Base Layer Name | Details | Description |
| --- | --- | --- |
| encode_same (input, filter, kernel, dropout = True, normalization = True) | Conv2D {W (3,3), S: (1,1)} leakyReLU () if (normalization = True): Batch Normalization() If (dropout = True) DropOut(0.5) | Creates an output layer of the same size as the input layer, with the specified depth (filter) |
| encode_half (input, filter, kernel, dropout = True, normalization = True) | Conv2D {W (3,3), S: (2,2)} leakyReLU () if (normalization = True): Batch Normalization() If (dropout = True) DropOut(0.5) | Creates an output layer of half the size (length/width) as the input layer, with the specified depth (filter) |
| encode_double (input, filter, kernel, dropout = True, normalization = True) | Conv2D Transpose{W (3,3), S: (2,2)} leakyReLU () if (normalization = True): Batch Normalization() If (dropout = True) DropOut(0.5) | Creates an output layer of double the size (length/width) as the input layer, with the specified depth (filter) |
| LL (input, filter, kernel, dropout = True, normalization = True) | Conv2D {W (3,3), S: (1,1)} ReLU () if (normalization = True): Batch Normalization() If (dropout = True) DropOut(0.5) | Creates an output layer of the same size as the input layer, with the specified depth (filter), using ReLU activation function |
| intermediate_enc_dec (input, filter) | p = encode_same (input, dwf) p = encode_half (input, dwf*4) p = encode_ half (input, dwf*16) | Combination of encode_same, encode_half and encode_double blocks to |
| | p = encode_same (input, dwf*64) p = encode_double (input, dwf*16) p = encode_double (input, dwf*4, F) p = encode_same (input, 1, F, F) | create an encoder-decoder like structure used in the network |

Input Encoder/Decoder (dwf = 4, F = False, T = True)

| Layer | Output |
| --- | --- |
| inp = encode_same (input,dwf,F,F) | 128 x 128 |
| d0 = encode_half (inp, dwf*4) | 64 x 64 |
| d1 = encode_half (d0, dwf*16) | 32 x 32 |
| d2 = encode_half (d1, dwf*64) | 16 x 16 |
| d3 = encode_half (d2, dwf*128) | 8 x 8 |
| d4 = encode_half (d3, dwf*256) | 4 x 4 |
| d5 = encode_double (d4, dwf*128) | 8 x 8 |
| d6 = encode_double (d4, dwf*64) | 16 x 16 |
| d7 = encode_same (d4, dwf*64) | 16 x 16 |
| d8 = encode_double (d4, dwf*16) | 32 x 32 |
| d9 = encode_double (d4, dwf*4) | 64 x 64 |
| d10 = encode_same (d4, dwf) | 64 x 64 |

DWT Layer (dwf = 4, F = False, T = True)

| Layer | Output |
| --- | --- |
| ca1, ch1, cv1, cd1 = slice along last axis (d10) | (64 x 64) |
| ll1 = LL (ca1, dwf) | 64x64 |
| ll1 = LL (ll1, dwf*4) | 64x64 |
| ll1 = LL (ll1, dwf*16) | 64x64 |
| ll1 = LL (ll1, dwf*32) | 64x64 |
| ll1 = encode_half (ll1, dwf*64) | 32 x 32 |
| ll1 = encode_same (ll1, dwf*16) | 32 x 32 |
| ll1 = encode_same (ll1, dwf*4) | 32 x 32 |
| ll1 = encode_same (ll1, dwf) | 32 x 32 |
| ca2, ch2, cv2, cd2 = slice along last axis (ll1) | (32 x 32) |
| ll2 = LL (ca2, dwf) | 32 x 32 |
| ll2 = LL (ll2, dwf*4) | 32 x 32 |
| ll2 = LL (ll2, dwf*16) | 32 x 32 |
| ll2 = LL (ll2, dwf*64) | 32 x 32 |
| ll2 = LL (ll2, dwf*16) | 32 x 32 |
| ll2 = LL (ll2, dwf*4, F) | 32 x 32 |
| ll2 = LL (ll2, 1, F, F) | 32 x 32 |
| hl2 = intermediate_enc_dec (ch2, 1) | 32 x 32 |
| lh2 = intermediate_enc_dec (cv2, 1) | 32 x 32 |
| hh2 = intermediate_enc_dec (cd2, 1) | 32 x 32 |
| hl1 = intermediate_enc_dec (ch1, 1) | 64 x 64 |
| lh1 = intermediate_enc_dec (cv1, 1) | 64 x 64 |
| hh1 = intermediate_enc_dec (cd1, 1) | 64 x 64 |
| ll2_1 = concatenate ([ll2, hl2], axis= 2) | 32 x 64 |
| ll2_2 = concatenate ([lh2, hh2], axis= 2) | 32 x 64 |
| ll1 = concatenate ([ll2_1, ll2_2], axis = 1) | 64 x 64 |
| a = concatenate ([ll1, hl1], axis = 2) | 64 x 128 |
| b = concatenate ([lh1, hh1], axis = 2) | 64 x 128 |
| op1 = (a, b, axis = 1) | 128 x 128 |

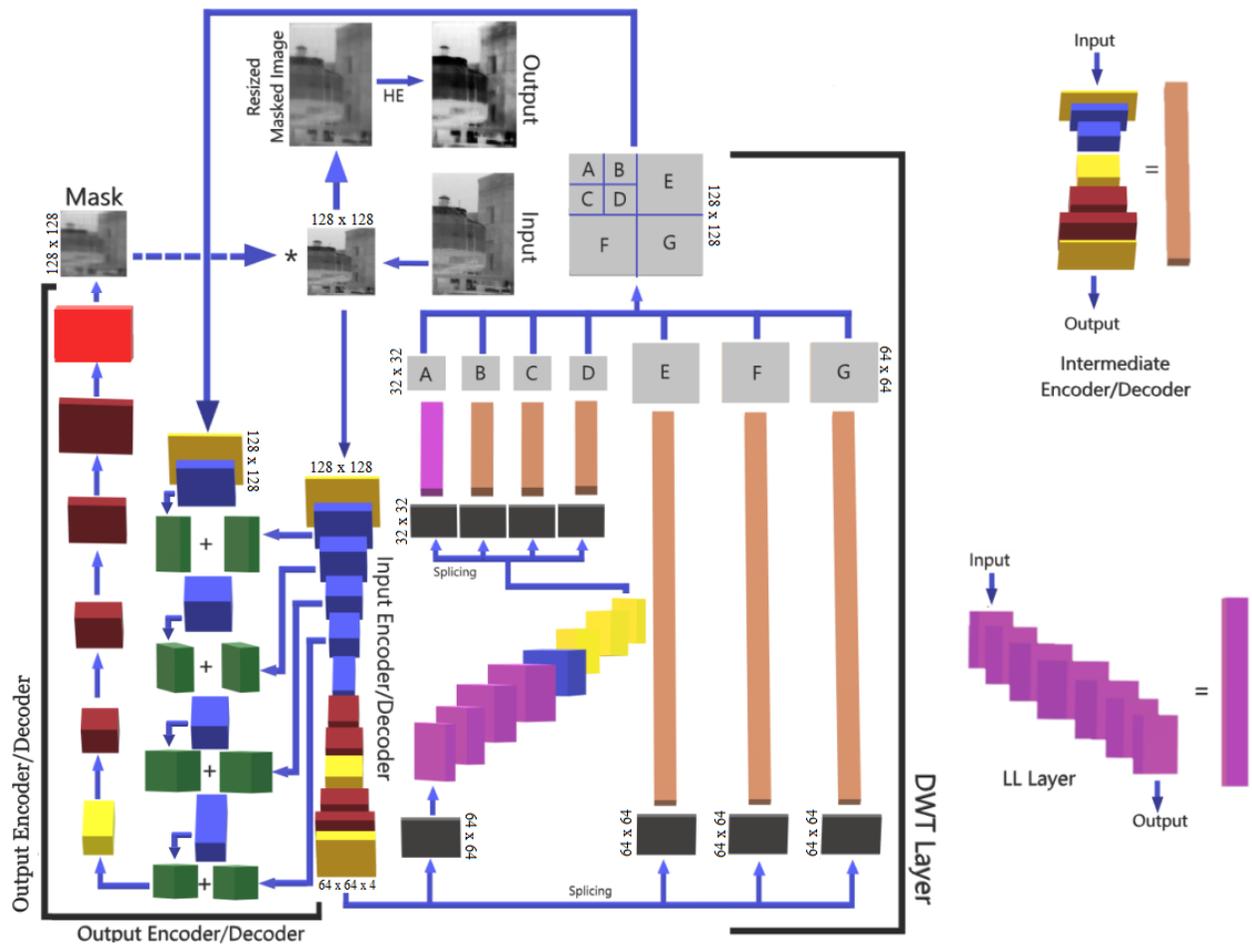

Fig. 1: The pink block represents an LL layer comprised of a combination of the individual encode_same blocks, the yellow ones are encode_same, blue represents encode_half and the red ones are encode_double. The intermediate encoder/decoder has been simplfied into a single orange block. The grey blocks are spliced single depth layers obtained by separating the 4 layers of an output block. The last red layer is a single depth Conv2D layer with sigmoid activation. The green blocks are concatenation blocks, which are joined along the last axis with the output from a previous layer and + represents the concatenation operation. The averaging is represented by the * operator and HE stands for Histogram Equalization. Detailed information on the model is available in Table 1

| op2 = encode_same (op1, dwf*4) | 128 x 128 |
|---|---|
| op2 = encode_half (op2, dwf*16) | 64 x 64 |
| op2 = concatenate ([op2, d0], axis = 3) | 64 x 64 |
| op2 = encode_half (op2, dwf*64) | 32 x 32 |
| op2 = concatenate ([op2,d1], axis = 3) | 32 x 32 |
| op2 = encode_half (op2, dwf*128) | 16 x 16 |
| op2 = concatenate ([op2,d2], axis = 3) | 16 x 16 |
| op2 = encode_half (op2, dwf*256) | 8 x 8 |
| op2 = concatenate ([op2,d3], axis = 3) | 8 x 8 |
| op2 = encode_same (op2, dwf*256) | 8 x 8 |
| op2 = encode_double (op2, dwf*128) | 16 x 16 |
| op2 = encode_double (op2, dwf*64) | 32 x 32 |
| op2 = encode_double (op2, dwf*16) | 64 x 64 |
| op2 = encode_double (op2, dwf*4, F) | 128 x 128 |
| output = Conv2D {W (3,3), S: (2,2)} | 128 x 128 |
| sigmoid () | |

In the above table all output layers have a depth equal to the number of filters used as the function arguments for convolution. Thus, for example, the layer *d0* has a shape of 64x64x16 when using an input shape of 128x128 in the *inp* layer. For slicing layers, outputs have a depth of 1 (i.e., it is a 2D matrix only).

The Input Encoder/Decoder is a basic encoder coupled with a decoder with Convolutional 2D Transpose layers (instead of statistical Upsampling layers) to preserve the gradient in between the layers. Also, we do not form a complete encoder-decoder, but put 1 less layer of the input dimension for feeding into the first level of our DWT layer.

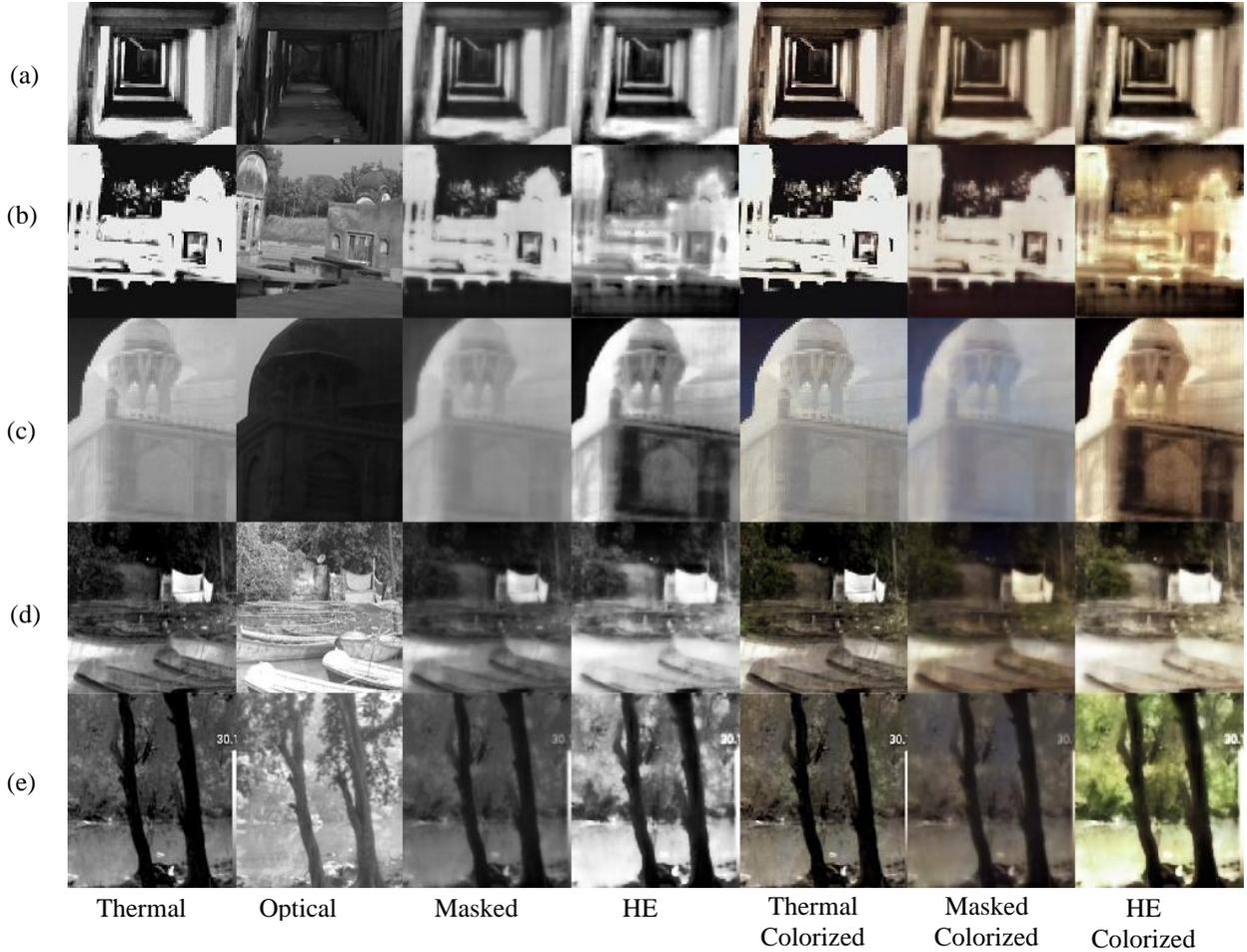

Fig. 2: A comparison of different images used for input versus the different outputs we obtain along with a blind testing done by colorizing the images with a network trained exclusively on optical images. HE represents Histogram Equalized masked images. All of the colorization is via the method in [18].

The Output Encoder/Decoder is also a general encoder decoder with a concatenation of the output from the input encoder with the output encoder. The concatenation is a layer-wise concatenation along the last axis. This is done so as to help the network learn the input texture in the output mask. Our method does not convert an input thermal image completely into a full optical mask, but instead minimizes the loss against a partial output image which has an averaged area embedded into the input thermal image. Thus, it would make sense to include this distribution into the output, which is what we try to encapsulate with the help of the skip connections. The output mask is obtained by using a last 2D Convolutional layer with the sigmoid activation which normalizes the output to (0,1) values.

We wanted to check if we could work in the Discrete Wavelet transform (DWT) domain instead of the usual normalized image domain for our fusion. The reasoning for this is that 2 Dimensional DWT (2DDWT) works on iteratively smaller scales of an image by halving the 2 axes of an input image, processing it and then reconstructing it back. In fact, this is similar to the logic of an encoder-decoder, except that an encoder-decoder based CNN works on the spatial domain and 2DDWT works on the frequency domain.

Since conversion from the spatial to the frequency domain is a standard signal processing algorithm, our logic was that a logically sound sufficiently complex deep network should be able to intuitively model it by itself. In fact, this is precisely the reason we alternate between 2 different blocks of deep network for modelling the LL blocks as opposed to the other (LH, HL, HH) blocks in our model. As can be noticed, for the LL blocks, we specifically use ReLU as the activation function, while we use LReLU for other blocks. In 2DDWT, LL blocks are confined to lie between 0 and a positive integer, which doubles in value with every successive iteration, which is the reason we use ReLU. When normalized, this essentially means that the values cannot go below 0, and are constrained between 0 and 1. On the other hand, the other blocks can have values which go below 0, and thus are effectively modelled by LReLU (where the output lower bound can have values below 0).

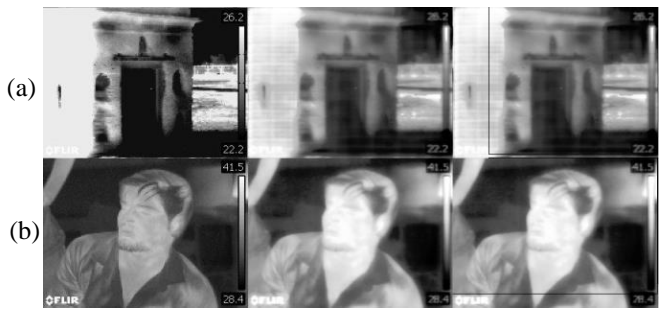
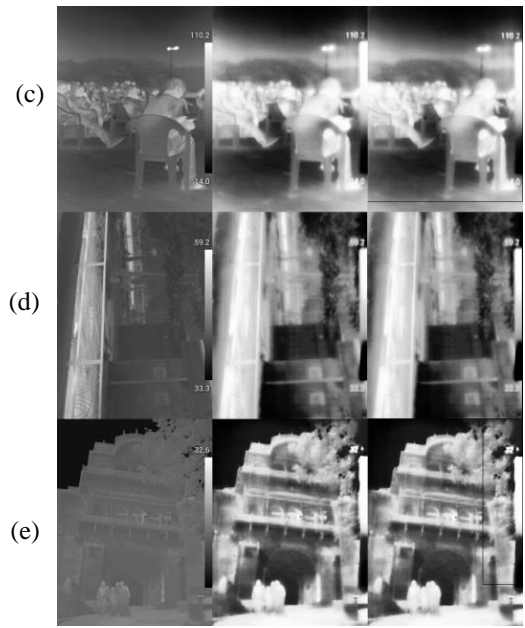

Fig. 3: A comparison of different images showing their RoF. (a) and (b) are images obtained via the FLIR E60 and (c), (d) and (e) are obtained via Sonel KT400 and have different image resolutions. The leftmost images are thermal inputs, the middle ones are Histogram Equalized masked outputs and the right most images are the outputs obtained after running the RoF algorithm on the images in the middle.

Of course, while the 1st and the 2nd level LL bands might be normalized to lie between 0 and 1, the 2nd level LH, HL and HH blocks may contain values below 0. However, these blocks combine to form the 1st level LL band, which is again normalized to lie between 0 and 1. This is exactly how we model our method as well.

It needs to be noted here that we design our model to represent only till the 2nd level decomposition. However, theoretically, one could go even deeper. We did not opt for this because of two reasons. Firstly, the size of our data was 128x128. We could not have data at the 256x256 size since the database we used had several images which had a maximum size of 240 in one dimension. Secondly, another level of decomposition would bring the output size of the patches down to 4x4, which would render our method unusable. Also, at one point, the increase in complexity would overrule the optimization of the loss.

As can be seen in our model, we decide to create a method that is able to create patches of localized data by creating an encoder-decoder structure for each scale of resolution we work on. The reason we decide to use this structure for the localized resolution paths is because we find that a localized encoder-decoder structure is able to lower the absolute loss by about 20%, as opposed to using patches with more depth at the same resolution. We also use skip connections between the 4 levels of the input encoder (before feeding it into our multi resolution kernels) and the output encoder (after we have obtained the final DWT kernel). We find that this optimizes the absolute loss by around 18% at the cost of about 10 times as many parameters. We wanted to find the optimum loss with skip connections between all the layers of the input encoder and the output encoder till a resolution of 1x1 was reached. However, that created an excessive overhead of parameters at 110 times the first variant, which did not fit in our hardware. So, we produced the model with only 4 localized encoder-decoder like structure as 6 skip connections did not give any significant change in accuracy over using 4, even at the cost of 20 times the initial number of parameters.

We decided to use Adaptive Moment Optimization (ADAM) as the optimizer because our model actually tries to optimize the loss function similar to what ADAM is doing as explained below.

The optimizer tries to lower the loss by changing values of $n^{th}$ moment, which is defined as

$$m_n = E[X^n] \qquad (1)$$

where $X$ represents the data and $E$ is the expectation. Since ADAM works by minimizing loss through moving average optimization, with the help of 2 constants ($\beta_1 = 0.9$ and $\beta_2 = 0.999$), the $k_{th}$ mini-batch moving averages at the $i^{th}$ level reduces to

$$\hat{m}_k = \frac{m_k}{1-\beta_1^i}, \hat{v}_k = \frac{v_k}{1-\beta_2^i} \qquad (2)$$

where $m$ and $v$ represent the moving averages.

Thus, we see that as the level goes deeper, the loss becomes lower and more local. The local convergence of ADAM optimizer has been relied on heavily for its choice of optimizer and has been already proven in [25]. This is what we are trying to achieve with our method as well, wherein the levels are logically represented by the parallel levels of DWT layer described in Table 1.

However, there is another hyper parameter, the loss function which forms an integral part of a deep network. We use *logcosh* as the loss for our current model. As stated in [1], if we consider wavelet based data, geodesic distance based on the Riemannian manifold is a good estimator as a distance measure. Since the Riemannian manifold is a part of the hyperbolic plane, we decided to use the *logcosh* loss measure representing the logarithm value of the hyperbolic cosine of the error predicted, given by

$$logcosh(x) = \begin{cases} \frac{x^2}{2} & for\ x \ll 1 \\ abs(x) - \log(2) & otherwise \end{cases} \quad (3)$$

Once we have the mask from our deep network, we fuse it with the thermal prior according the simple averaging rule:

$$O_i = (T_i + M_i)/2 \quad (4)$$

where $O_i$ represents the averaged output image pixel, $T_i$ is corresponding the thermal prior pixel and $M_i$ is the equivalent mask pixel for each $i^{th}$ pixel. We are trying to obtain an image that already has the thermal image as a part of the output.

Finally, we go on to equalize the fused image in order to obtain a final output image which has a better distribution of illumination for better visibility. Here we point out that there is no meaning to equalizing a thermal image. This is because thermal images are already histogram equalized by the capturing device since thermal images use all 256 levels of illumination. This is evident from the thermal bar present on the right side of thermal images. We present a comparison of the thermal input, the fused image and the final output in Fig. 2.

### 3.2. (RoF)

When looking at research works focused on fusion, we have found that there is no objective measure which could provide a bounding box for the localized regions of fusion. This becomes relevant in works such as this one, where we are focusing on regions which should have localized content for fusion.

Hence, we propose a measure called Region of Fusion (RoF), based on localized Region of Interest (ROI), which can be objectively calculated given a fused image and the input from which it is obtained. This method is fully customizable in regards to the distance metric that is used to calculate the region similarity and can be used on fusion methods which are either DL or statistically based. It has a low computational complexity on par with the size of the image (constrained by the similarity measure being used).

The idea behind the method is to take a score for the variation of full image between the thermal and the fused output and then calculate the area of both. We iteratively reduce the size of the region by 1 and check the percentage reduction of variation with the percentage reduction in area. If variation is less than the change in area, we stop and define the region as the final region. The full algorithm is discussed in details as Algorithm 1 below.

---
Algorithm 1
---
**Input**: Thermal-Fused grayscale image pair
**Output**: Fused Image with a RoF
1. for each image pair in list:
2. obtain image size as $x = 0$ to $m$, $y = 0$ to $n$
    2.1. check = 0
    2.2. while $x<n$, with $x = 0$, $y = n$
        2.2.1. if check = 1
            2.2.1.1. $x = x+1$
            2.2.1.2. continue
        2.2.2. *calculate the measure of similarity between the thermal and the fused image patch for region ($x = 0$ to $m$, $y = 0$ to $n$) as $M_1$*
        2.2.3. *calculate the measure of similarity between the thermal and the fused image patch for region ($x = 1$ to $m$, $y = 0$ to $n$) as $M_2$*
        2.2.4. calculate the area $A_1 = (m-x)*n$ and $A_2 = (m-(x+1))*n$
        2.2.5. if $M_1 / M_2 < A_1 / A_2$
            2.2.5.1. $x = x+1$
            2.2.5.2. $X_1 = x$
        2.2.6. *else*
            2.2.6.1. *check = 1*
    2.3. *check = 0*
    2.4. *while $x>X_1$, with $x = m$, $y = n$*
        2.4.1. if check = 1
            2.4.1.1. $x = x-1$
            2.4.1.2. continue
        2.4.2. *calculate the measure of similarity between the thermal and the fused image patch for region ($x = X_1$ to $m$, $y = 0$ to $n$) as $M_1$*
        2.4.3. *calculate the measure of similarity between the thermal and the fused image patch for region ($x = X_1$ to $m-1$, $y = 0$ to $n$) as $M_2$*
        2.4.4. calculate the area $A_1 = (m- X_1)*n$ and $A_2 = ((m-1) - X_1 )*n$
        2.4.5. if $M_1 / M_2 < A_1 / A_2$
            2.4.5.1. $x = x-1$
            2.4.5.2. $X_2 = x$
        2.4.6. *else*
            2.4.6.1. *check = 1*
    2.5. *repeat steps 2.1 – 2.2 keeping x as constant from $X_1$ to $X_2$ and varying y from 0 to n to get $Y_1$*
    2.6. *repeat steps 2.3 – 2.4 keeping x as constant from $X_1$ to $X_2$ and varying y from n to $Y_1$ to get $Y_2$*
    2.7. *draw a box on the fused image from ($X_1$, $Y_1$) to ($X_2$, $Y_2$) to get the RoF*

---

As can be understood from the above algorithm, the measure for similarity (or dissimilarity) can be changed as needed. We have used the sum of the square of difference of pixel values as a measure of dissimilarity in our case, but one can use other scores, like Structural Similarity Index Measure (SSIM). Also, since this is an unbiased score, one can opt to combine existing DL based fusion methods like [11, 12] with our score to provide a better measure of the fused region to focus on the output.

One may note that while RoF can model any fusion based distribution for a region of interest, the data distribution needs to be on the same scale. For example, if we work with data distributions modelled on Wavelet transformed data (without converting it back into an image), which are multi resolution data distributions containing several scales of an image, it would not work. This is because RoF is designed to work on data in a single scale only by determining a local maximum for the bounding box.

### 3.3. Database

We use the thermal-visual paired images dataset [6] for our work. It was presented as a part of a work on colorizing thermal images. We use 10 random cuts of the input images, while keeping the annotated region inside the cut. This limits our data, and we are able to create only 89,442 pairs of images for our experiment. This is because while it is possible to take a random cut from an input pair if the objective was just the production of a fused thermal-optical image, we try to create a model which is able to create a localized region for the final output instead of a uniform fused image. Examples of these images are shown in the Supplementary Section 2.

The database we propose has 1873 thermal images hand annotated by us. The annotation in rectangular bounding boxes is done using the tool VGG Image Annotator (VIA) [7]. We annotate the images into 5 different classes, Nature (nat), Animal (ani), Human (hum), Crowd (cro) and Modern Infrastructure (inf). Each image may have multiple annotated objects inside it, which is how we are able to obtain more than one sub-image for each individual annotated image. A few of these are included in the Supplementary Section 1.

However, the database [6] also contained the paired visual equivalent for each of these thermal images, which has lent a way to further augment our dataset. We applied an optical Region of Interest bounding box algorithm on the optical images to create additional data. The logic behind this is that since we are proposing a localized fusion method, the fusion should occur from both directions, and not just from thermal to optical. There might be objects present in the optical domain which are not well visible in the thermal domain (objects at the same thermal profile range). We came across different object identification algorithms like [8, 9] etc, but most of them were either image classifiers only or did not provide multiple bounding boxes over a single image, as we required. Moreover, since the database we were using as our background base was focused on multiple classes, not of a very high resolution and of different sizes, we wanted to find a low overhead high accuracy algorithm which would be able to fit in our case. Thus, we decided to opt for DEtection TRansformer (DETR) [10] for our use. DETR is a state of the art localized low cost object annotator which has multiple object classes, trained on optical images. We use the public code that they provide, and obtain object annotations on the optical images in the database and transpose these boxes on their thermal counterparts to finally obtain the localized database we are using for our use. Of course, since we have only 5 classes in our annotation, we simplify the annotation provided by DETR into our annotation labels by changing classes like bus, car, laptop, truck etc into Modern Infrastructure, sheep, horse etc into animal and so on.

The final database we propose has 5 different classes labeled as a number from 1 to 5. Once this is done, we take 10 random cuts around the annotated region, for each of the images with individual annotations constraining the minimum size for each image to be 128 x 128 and combine the thermal and optical information in the annotated region following Eq. (4). We finally obtain 89,442 image pairs, which we use in our work.

It should be noted that since there is no extra restriction on the annotated region size, our final database for the model comprises of images of widely differing sizes, which we normalize to 128x128 keeping parity in the input and output image sizes in the deep network. We reshape it back to the original image size after we obtain the mask to create the final masked output for equalization and obtain the final output.

### 4. Experimental Results

We use 3 different objective scores to evaluate our method. Since we did not find any thermal image fusion evaluation score, we choose Structural Similarity Index Measure (SSIM) [20], Cosine-similarity (Cossim) and Mean Square Error (MSE) for our evaluation. These scores are denoted in Table 2.

|        | Thermal vs output | Thermal vs visual | Visual vs output |
|--------|-------------------|-------------------|------------------|
| SSIM   | **0.873**         | 0.2641            | **0.307**        |
| Cossim | **0.932**         | 0.889             | **0.916**        |
| MSE    | **117.91**        | 9613.741          | **8620.87**      |

Table 2: Objective score comparison between thermal images, optical images and the masked average outputs

In Table 2, we show a comparison between our masked average outputs versus their thermal and optical counterparts. We use 3 different measures of similarity. The first column shows how similar our averaged output is to the thermal images. Similarly the third column shows their similarity with the visual counterparts. The middle one is the similarity of the thermal images to their optical counterparts and provides the baseline against which we compare our values. Of course, the scores between the thermal and the averaged output is much better than the ones between the optical and the average output, because we fuse the mask with the thermal image before comparison.

## 5. Discussion

Since there is no direct method of comparison for showing that our method produces a significantly different output (as all of it is in grayscale domain and optical features incorporated in the mask are not immediately identifiable), we opt for an indirect method to show this. We use a neutral testing method of coloring of the thermal image and the final output we are producing. The coloring is done via the method explained in [18] for optical image colorization. We use the online demo method they provide for the same, without training it on our database for the blind testing.

As can be seen in Fig 2, the Histogram Equalized (HE) images contain more texture as compared to the thermal images. This is especially noticeable in image (e), (a) and (c). We include (a) since it has a noticeably bad optical image since the setting for the photograph possessed a bad illumination. Similarly, (b) has a binary thermal image. These kinds of images occur when the levels of the thermal imager for the upper and lower limits of temperature to be captured are very near the limits of temperature of the surrounding. As can be seen, both (a) and (b) provide outputs which have noticeably improved texture in the masked images. The colored mask as well as the colorized HE masked images show the same. In case of (c), we see that the walls of the structure in the image have a lower temperature. This region becomes prominent in the HE masked images. Finally, we note that the images for both (d) and (e) are quite different from the input thermal images. This is especially noticeable in all of these images when we consider the blind testing method we provide, in which we color each of these images via the method described in [18], which is a pure optical coloring technique. Our method shows that the color improves as compared to the thermal images. Of course, this is possible because the temperature profile is not well segmented in most thermal images, which is why these are different from optical images. However, if there exist thermal images which have very well defined levels of separation in between objects, our method would not perform as well since the texture might be interpreted as noise by a machine learning method. None of the images shown in Fig. 2 are of the same size because all of them were random cuts from thermal images.

We show the results from our measure of fusion, RoF in Fig. 3. The images we use here are those which are published in [6] as being unregistered. Thus, we did not have the optical counterparts of these images, and they were not used in training our DL algorithm. We obtain the HE masked images for each of the thermal inputs and then run our RoF identifying algorithm on them. The texture difference is relevant in case of images (a), (b) and (c), where we see a clear region of interest. In case of (d), the region is around almost the full image. However, in (e), the region is quite outside the expected region of interest. This is because, in (e), we see that the thermal image has well defined levels of separation for regions. However, our method does detect a region where the score varies enough to make a RoF. In images such as this, since the thermal image itself is well segmented and possess well defined visual features, we would not opt for a fusion method. However, we include this result to show that this case may also occur.

We use 89,352 images for training and 90 images for validation. We finally use 294 images for testing against paired images, which are random cuts from registered images in [6] and 438 images for testing on a blind dataset, images that were unregistered in the dataset.

All experiments have been conducted on Keras 2.2.4, with Tensorflow 1.13.1 as the backend, using a 1080Ti GPU on a 7820X i7 chipset processor. This work was supported by the Computer Visions and Biometrics Lab (CVBL), IIIT-Allahabad.

## 6. Conclusion

We present a novel method demonstrating that it is possible to fuse thermal images with optical priors having annotated regions for focusing on specific regions. The model is both unique in its scope of work and the theoretical basis, wherein we show that the calculations are based on a separate logical space, constructed on the principles of 2 Dimensional Discrete Wavelet Transform. We also introduce a simple statistical score for identifying regions with significantly different distribution in output fused images. Lastly, we introduce a unique database [19] containing annotated thermal images on varying classes as a part of this work for public use.

While the outputs are promising, further scope lies in checking how the method behaves under the use of other metrics for loss optimization, like geodesic distance, what is the behavior of the method, when we use deeper networks and how to extract and process information in the joint domain for better processing of fused images.